\documentclass[letterpaper, 10 pt, conference]{ieeeconf}

\IEEEoverridecommandlockouts                              
\usepackage{graphics} 
\usepackage[colorlinks=true,allcolors=black]{hyperref}
\usepackage{stmaryrd} 
\usepackage{epsfig} 

\usepackage{times} 
\usepackage{amsmath} 
\usepackage{amssymb}  
\usepackage{float}
\usepackage{hyperref}
\usepackage{subcaption}
\usepackage{physics}
\usepackage{tikz}
\usepackage{pgfplots}
\usepackage{balance}
\usepackage{tabu}
\usepackage{multirow,booktabs}
\usepackage{cite}
\usepackage[linesnumbered,ruled,vlined]{algorithm2e}

\newcommand{\sref}[1]{{Section \ref{#1}}}
\newcommand{\fref}[1]{Fig. \ref{#1}}
\newcommand{\tref}[1]{Table \ref{#1}}

\title{\LARGE \bf
Multi-modal Semantic SLAM for Complex Dynamic Environments}

\author{Han Wang*, Jing Ying Ko* and Lihua Xie, \textit{Fellow}, IEEE

\thanks{*Jing Ying Ko and Han Wang contribute equally to this paper and are considered as jointly first authors.}%
\thanks{
The research is supported by the National Research Foundation, Singapore under its Medium Sized Center for Advanced Robotics Technology Innovation. 
}
\thanks{Jing Ying Ko, Han Wang and Lihua Xie are with the School of Electrical and Electronic Engineering,
Nanyang Technological University, 50 Nanyang Avenue, Singapore 639798.
        {\tt\small e-mail: \{hwang027, E170043\}@e.ntu.edu.sg; elhxie@ntu.edu.sg}}
}

\begin{document}
 
\maketitle
\thispagestyle{empty}
\pagestyle{empty}

\begin{abstract}

Simultaneous Localization and Mapping (SLAM) is one of the most essential techniques in many real-world robotic applications.
The assumption of static environments is common in most SLAM algorithms, which however, is not the case for most applications.
Recent work on semantic SLAM aims to understand the objects in an environment and distinguish dynamic information from a scene context by performing image-based segmentation. However, the segmentation results are often imperfect or incomplete, which can subsequently reduce the quality of mapping and the accuracy of localization. 
In this paper, we present a robust multi-modal semantic framework to solve the SLAM problem in complex and highly dynamic environments. We propose to learn a more powerful object feature representation and deploy the mechanism of looking and thinking twice to the backbone network, which leads to a better recognition result to our baseline instance segmentation model. Moreover, both geometric-only clustering and visual semantic information are combined to reduce the effect of segmentation error due to small-scale objects, occlusion and motion blur. 
Thorough experiments have been conducted to evaluate the performance of the proposed method. The results show that our method can precisely identify dynamic objects under recognition imperfection and motion blur. Moreover, the proposed SLAM framework is able to efficiently build a static dense map at a processing rate of more than 10 Hz, which can be implemented in many practical applications.
Both training data and the proposed method is open sourced\footnote{\url{https://github.com/wh200720041/MMS_SLAM}}. 

\end{abstract}
\section{INTRODUCTION}
Simultaneous Localization and Mapping (SLAM) is one of the most significant capabilities in many robot applications such as self-driving cars, unmanned aerial vehicles, etc. 
Over the past few decades, SLAM algorithms have been extensively studied in both Visual SLAM such as ORB-SLAM \cite{mur2017orb} and LiDAR-based SLAM such as LOAM \cite{zhang2014loam} and LeGO-LOAM \cite{shan2018lego}.
Unfortunately, many existing SLAM algorithms assume the environment to be static, and cannot handle dynamic environments well.
The localization is often achieved via visual or geometric features such as feature points, lines and planes without including semantic information to represent the surrounding environment, which can only work well under static environments. 
However, the real-world is generally complex and dynamic. In the presence of moving objects, pose estimation might suffer from drifting, which may cause the system failure if there are wrong correspondences or insufficient matching features \cite{tan2013robust}. 
The presence of dynamic objects can greatly degrade the accuracy of localization and the reliability of the mapping during the SLAM process. 


\begin{figure}[t]
\begin{center}
\vspace{5pt}
\includegraphics[width=0.99\linewidth]{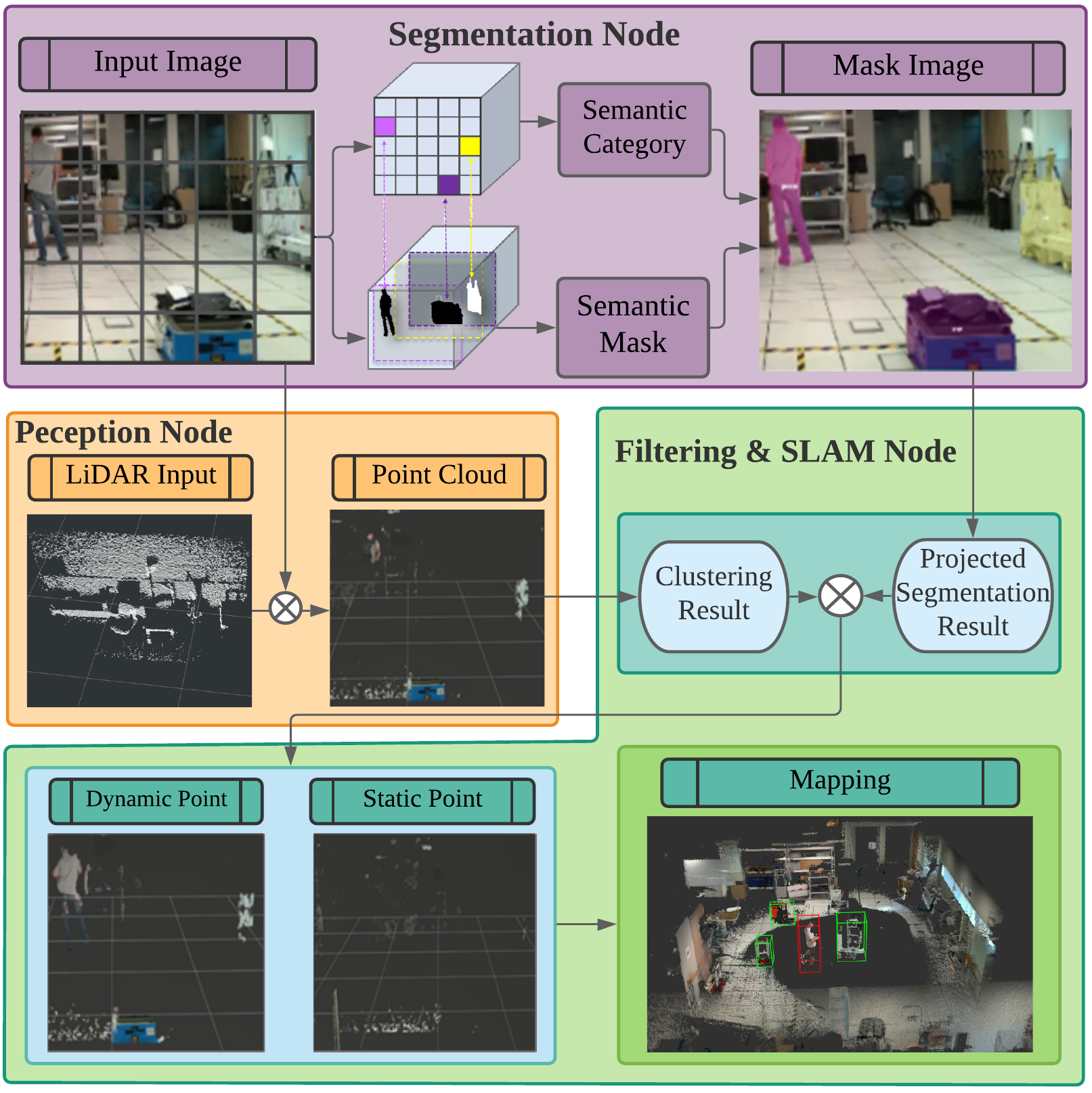}
\captionsetup{justification=justified}
\caption{System overview of the proposed multi-modal semantic SLAM. Compared to traditional semantic SLAM, we propose to use multi-modal method to improve the efficiency and accuracy of the existing SLAM methods in the complex and dynamic environment. Our method significantly reduces the localization drifts caused by dynamic objects and performs dense semantic mapping in real time.}
\label{fig: title_graph}
\end{center}
\vspace{-15pt}
\end{figure}



Advancements in deep learning have enabled the developments of various instance segmentation networks based on 2D images \cite{he2017mask, bolya2019yolact}. Most existing semantic SLAMs leverage the success of deep learning-based image segmentation, e.g., dynamic-SLAM \cite{xiao2019dynamic} and DS-SLAM \cite{yu2018ds}. However, the segmentation results are not ideal under dynamic environments. Various factors such as small-scale objects, objects under occlusion and motion blur contribute to challenges in 2D instance segmentation. For example, the object is partially recognized under motion blur or when it is near to the border of the image. These can degrade the accuracy of localization and the reliability of the mapping. 
Some recent works target to perform deep learning on 3D point clouds to achieve semantic recognition \cite{lei2020occuseg, li2020pointgroup}. However, 3D point cloud instance segmentation does not perform as well as its 2D counterpart due to its smaller scale of training data and high computational cost. There are several reasons: 1) 3D point cloud instance segmentation such as PointGroup takes a long computation time (491ms) \cite{jiang2020pointgroup}; 2) it is much less efficient to label a point cloud since the geometric information is not as straightforward as the visual information; 3) it is inevitable to change the viewpoint in order to label a point cloud \cite{behley2019iccv}, which increases the labeling time.  

In this paper, we propose a robust and computationally efficient multi-modal semantic SLAM framework to tackle the limitation of existing SLAM methods in dynamic environments. 
We modify the existing backbone network to learn a more powerful object feature representation and deploy the mechanism of looking and thinking twice to the backbone network, which leads to a better recognition result to our baseline instance segmentation model. Moreover, we combine the geometric-only clustering and visual semantic information to reduce the effect of motion blur. Eventually the multi-modal semantic recognition is integrated into the SLAM framework which is able to provide real-time localization in different dynamic environments. The experiment results show that the segmentation errors due to misclassification, small-scale object and occlusion can be well-solved with our proposed method. The main contributions of this paper are summarized as follows:

\begin{itemize}
\item We propose a robust and fast multi-modal semantic SLAM framework that targets to solve the SLAM problem in complex and dynamic environments. Specifically, we combine the geometric-only clustering and visual semantic information to reduce the effect of segmentation error due to small-scale objects, occlusion and motion blur.
\item We propose to learn a more powerful object feature representation and deploy the mechanism of looking and thinking twice to the backbone network, which leads to a better recognition result to our baseline instance segmentation model.  
\item A thorough evaluation on the proposed method is presented. The results show that our method is able to provide reliable localization and a semantic dense map. 
\end{itemize}

The rest of the paper is organized as follows: \sref{sec:related-work} presents an overview of the related works regarding the three main SLAM methods in dynamic environments. \sref{sec:methodology} describes the details of the proposed SLAM framework. \sref{sec:experiment} provides quantitative and qualitative experimental results in dynamic environments. \sref{sec:conclusion} concludes this paper.

\begin{figure*}[t]
\begin{center}

\includegraphics[width=0.99\linewidth]{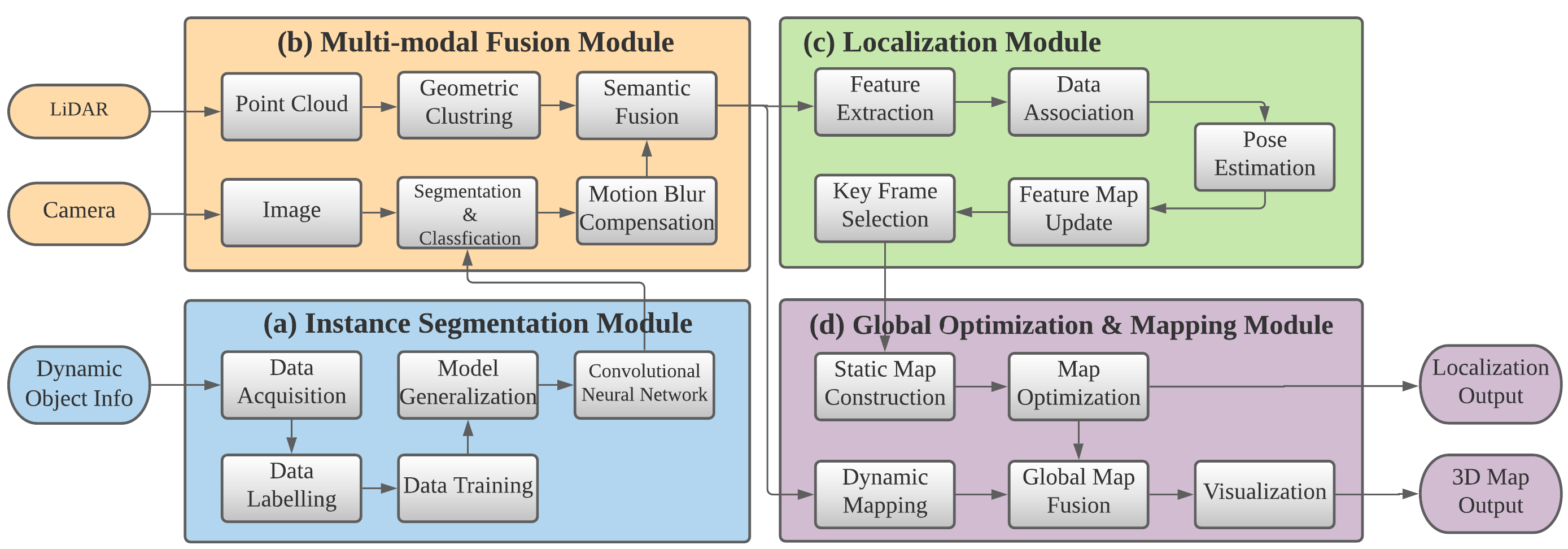}
\captionsetup{justification=justified}
\caption{Flow chart of the proposed method. Our system consists of four modules: (a) semantic fusion module; (b) semantic learning module; (c) localization module; (d) global optimization and mapping module.}
\label{fig: flowchart}
\vspace{-8pt}
\end{center}
\end{figure*}

\section{RELATED WORK}\label{sec:related-work}

In this section, we present the existing works that address SLAM problems in dynamic environments. The existing dynamic SLAM can be categorized into three main methods: feature consistency verification method, deep learning-based method and multi-modal-based method.

\subsection{Feature Consistency Verification}\label{AA}
Dai et al. \cite{dai2020rgbd} presents a segmentation method using the correlation between points to distinguish moving objects from the stationary scene, which has a low computational requirement. Lee et al. \cite{li2017rgbd} introduces a real-time depth edge-based RGB-D SLAM system to deal with a dynamic environment. Static weighting method is proposed to measure the likelihood of the edge point being part of the static environment and is further used for the registration of the frame-to-keyframe point cloud. These methods generally can achieve real-time implementation without increasing the computational complexity. Additionally, they need no prior knowledge about the dynamic objects. However, they are unable to continuously track potential dynamic objects, e.g., a person that stops at a location temporarily between moves is considered as a static object in their work.

\subsection{Deep Learning-Based Dynamic SLAM}
Deep learning-based dynamic SLAM usually performs better than feature consistency verification as it provides conceptual knowledge of the surrounding environment to perform the SLAM tasks. Xun et al. \cite{xun2021sad} proposes a feature-based visual SLAM algorithm based on ORB-SLAM2, where a front-end semantic segmentation network is introduced to filter out dynamic feature points and subsequently fine-tune the camera pose estimation, thus making the tracking algorithm more robust. Reference \cite{yu2018dsslam} combines a semantic segmentation network with a moving consistency check method to reduce the impact of dynamic objects and generate a dense semantic octree map. A visual SLAM system proposed by \cite{bescos2018dynaslam} develops a dynamic object detector with multi-view geometry and background inpainting, which aims to estimate a static map and reuses it in long term applications. However, Mask R-CNN is considered as computationally intensive; as a result, the whole framework can only be performed offline.

Deep learning-based LiDAR SLAM in dynamic environments are relatively less popular than visual SLAM. Reference \cite{chen2019suma} integrates semantic information by using a fully convolutional neural network to embed these labels into a dense surfel-based map representation. However, the adopted segmentation network is based on 3D point clouds, which is less effective as compared to 2D segmentation networks. Reference \cite{zhao2019liom} develops a laser-inertial odometry and mapping method which consists of four sequential modules to perform a real-time and robust pose estimation for large scale high-way environments. Reference \cite{jian2019semantic} presents a dynamic objects-free LOAM system by overlapping segmented images into LiDAR scans. Although deep learning-based methods can effectively alleviate the impact of dynamic objects on the SLAM performance, they are normally difficult to operate in real-time due to the implementation of deep-learning neural networks which possess high computational complexity.

\subsection{Multi-modal-based Dynamic SLAM}
Multi-modal approaches are also explored to deal with dynamic environments.  Reference \cite{jeong2018toward} introduces a multi-modal sensor-based semantic mapping algorithm to improve the semantic 3D map in large-scale as well as in featureless environments. Although this work is similar to our proposed method, it incurs higher computational cost as compared to our proposed method. A LiDAR-camera SLAM system \cite{jiang2016static} is presented by applying a sparse subspace clustering-based motion segmentation method to build a static map in dynamic environments. Reference \cite{zhang2012sensor} incorporates the information of a monocular camera and a laser range finder to remove the feature outliers related to dynamic objects. However, both reference \cite{jiang2016static} and \cite{zhang2012sensor} can only work well in low dynamic environments. 
%

\section{METHODOLOGY}\label{sec:methodology}
In this section, the proposed method will be discussed in detail. \fref{fig: flowchart} illustrates an overview of our framework.
It is mainly composed of four modules, namely instance segmentation module, multi-modal fusion module, localization module and global optimization \& mapping module. Instance segmentation module uses a real-time instance segmentation network to extract the semantic information of all potential dynamic objects that are present in an RGB image. The convolution neural network is trained offline and is later implemented online to achieve real-time performance. Concurrently, the multi-modal fusion module transfers relevant semantic data to LiDAR through sensor fusion and subsequently uses the multi-modal information to further strengthen the segmentation results. The static information is used in the localization module to find the robot pose, while both static information and dynamic information are utilized in the global optimization and mapping module to build a 3D dense semantic map. 

\subsection{Instance Segmentation \& Semantic Learning}
A recent 2D instance segmentation framework \cite{wang2020solo} is employed in our work due to its ability to outperform other state-of-the-art instance segmentation models, both in segmentation accuracy and inference speed. 
Given an input image $\displaystyle I$, our adopted instance segmentation network predicts a set of \{$\displaystyle C\textsubscript{i}, M\textsubscript{i}\}^{\text{n}}_{\text{i=1}}$, where $\displaystyle C\textsubscript{i}$ is a class label and $\displaystyle M\textsubscript{i}$ is a binary mask, $\displaystyle n$ is the number of instances in the image. The image is spatially separated into $\displaystyle N \times \displaystyle N$ grid cells. If the center of an object falls into a grid cell, that grid cell is responsible for predicting the semantic category $\displaystyle C\textsubscript{ij}$ and semantic mask $\displaystyle M\textsubscript{ij}$ of the object in category branch $B\textsubscript{c}$ and mask branch $\displaystyle P\textsubscript{m}$ respectively:
\begin{subequations}
\begin{align}
\displaystyle B\textsubscript{c}(\displaystyle I, \theta\textsubscript{c})&: I \rightarrow{\{\displaystyle C\textsubscript{ij} \in \mathbb{R}^{\lambda} \mid \displaystyle i,j = 0, 1, ..., N\}},\\
\displaystyle P\textsubscript{m}(\displaystyle I, \theta\textsubscript{m})&: I \rightarrow{\{\displaystyle M\textsubscript{ij} \in \mathbb{R}^{\phi} \mid \displaystyle i,j = 0, 1, ..., N\}},
\end{align}
\end{subequations}
where $\theta\textsubscript{c}$ and $\theta\textsubscript{m}$ are the parameters of category branch $\displaystyle B\textsubscript{c}$ and mask branch $\displaystyle P\textsubscript{m}$ respectively. $\lambda$ is the number of classes. $\phi$ is the total number of grid cells. The category branch and mask branch are implemented with a Fully Connected Network (FCN). $\displaystyle C\textsubscript{ij}$ has a total of $\lambda$ elements. Each element of $\displaystyle C\textsubscript{ij}$ indicates the class probability for each object instance at grid cell $\displaystyle (i, j)$. In parallel with the category branch, $\displaystyle M\textsubscript{ij}$ has a total of $\displaystyle N\textsuperscript{2}$ elements \cite{wang2020solo}. Each positive grid cell $\displaystyle (i, j)$ will generate the corresponding instance mask in $\displaystyle k\textsuperscript{th}$ element, where  $\displaystyle k\textsuperscript{th} = \displaystyle i \cdot N + j$.
\begin{figure*}[t]
\begin{center}
\vspace{4pt}
\includegraphics[width=0.79\linewidth]{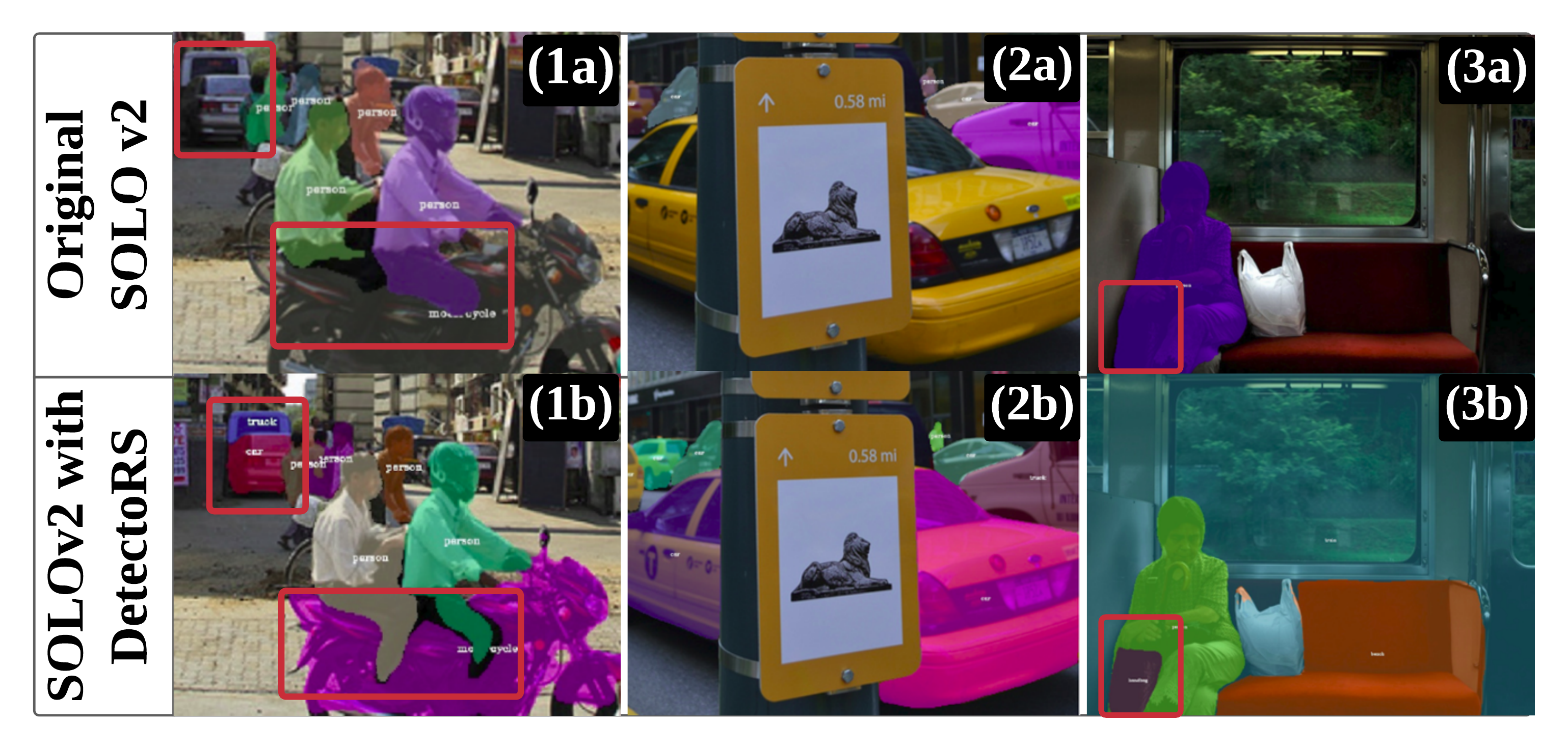}
\captionsetup{justification=justified}
\caption{Comparison of the original SOLOv2 with the proposed method. Our segmentation results achieve higher accuracy: In (1b), our method can preserve a more detailed mask for the rider on a motorcycle compared to the SOLOv2 result in (1a); In (2b), we can handle the occluded object while it is not detected in (2a); In (3b), our method can accurately predict the mask for a handbag compared to (3a).
}
\label{fig: compare}
\end{center}
\vspace{-8pt}
\end{figure*}
Since our proposed SLAM system is intentionally designed for real-world robotics applications, computational cost for performing instance segmentation is our primary concern. Therefore, we use a light-weight version of SOLOv2 with lower accuracy to achieve real-time instance segmentation. To improve the segmentation accuracy, several methods have been implemented to build a more effective and robust feature representation discriminator in the backbone network. Firstly, we modify our backbone architecture from the original Feature Pyramid Network (FPN) to Recursive Feature Pyramid Network (RFP) \cite{qiao2020detectors}. Theoretically, RFP instills the idea of looking twice or more by integrating additional feedback from FPN into bottom-up backbone layers. This recursively strengthens the existing FPN and provides increasingly stronger feature representations. By offsetting richer information with small receptive field in the lower-level feature maps, we are able to improve the segmentation performance on small objects. Meanwhile, the ability of RFP to adaptively strengthen and suppress neuron activation enables the instance segmentation network to handle occluded objects more efficiently.
On the other hand, we replace the convolutional layers in the backbone architecture with Switchable Atrous Convolution (SAC). SAC operates as a soft switch function, which is used to collect the outputs of convolutional computation with different atrous rates. Therefore, we are able to learn the optimal coefficient from SAC and can adaptively select the size of receptive field. This allows SOLOv2 to efficiently extract important spatial information.  

The outputs are pixel-wise instance masks for each dynamic object, as well as their corresponding bounding box and class type. To better integrate the dynamic information to the SLAM algorithm, the output binary mask is transformed into a single image containing all pixel-wise instance masks in the scene. The pixel with the mask falling onto it is considered as “dynamic state” and otherwise is considered as “static state”. The binary mask is then applied to the semantic fusion module to generate a 3D dynamic mask. 

\subsection{Multi-Modal Fusion}
\subsubsection{Motion Blur Compensation}
The instance segmentation has achieved good performance on the public dataset such as the COCO dataset and the Object365 dataset \cite{wang2020solo,ghiasi2021simple}. However, in practice the target may be partially recognized or incomplete due to the motion blur on moving objects, resulting in ambiguous boundaries of a moving object. Moreover, motion blur effect is further enlarged when projecting the 2D pixel-wise semantic mask for a dynamic object to 3D semantic label, leading to point misalignment and inconsistency of feature point extraction. In the experiments, we find that the ambitious boundaries of dynamic targets will degrade the localization accuracy and produce noise when performing a mapping task. Therefore, we firstly implement morphological dilation to convolute the 2D pixel-wise mask image with a structuring element, for gradually expanding the boundaries of regions for the dynamic objects. The morphological dilation result marks the ambiguous boundaries around the dynamic objects. We take the both dynamic objects and their boundaries as the dynamic information, which will be further refined in the multi-modal fusion section.
\subsubsection{Geometric Clustering \& Semantic Fusion}
Compensation via connectivity analysis on Euclidean space \cite{rusu2010semantic} is also implemented in our work. Instance segmentation network has excellent recognition capability in most practical situations, however motion blur limits the segmentation performance due to ambiguous pixels between regions, leading to undesirable segmentation error. Therefore, we combine both point cloud clustering results and segmentation results to better refine the dynamic objects. In particular, we perform the connectivity analysis on the geometry information and merge with vision-based segmentation results. 

A raw LiDAR scan often contains tens of thousands of points. To increase the efficiency of our work, 3D point cloud is firstly downsized to reduce the scale of data and used as the input for point cloud clustering. Then the instance segmentation results are projected to the point cloud coordinate to label each point. The point cloud cluster will be considered as a dynamic cluster when most points (90\%) are dynamic labelled points. The static point will be re-labeled to the dynamic tag when it is close to a dynamic point cluster. And the dynamic point will be re-labelled when there is no dynamic points cluster nearby.

\subsection{Localization \& Pose Estimation}
\subsubsection{Feature Extraction} After applying multi-modal dynamic segmentation, the point cloud is divided into a dynamic point cloud $\mathcal{P}_{\mathcal{D}}$ and a static point cloud $\mathcal{P}_{\mathcal{S}}$. The static point cloud is subsequently used for the localization and mapping module based on our previous work \cite{wang2021light}. Compared to the existing SLAM approach such as LOAM \cite{zhang2014loam}, the proposed framework in \cite{wang2021light} is able to support real-time performance at 30 Hz which is a few times faster. It is also resistant to illumination variation compared to visual SLAMs such as ORB-SLAM \cite{mur2017orb} and VINS-MONO \cite{qin2018vins}. For each static point $\mathbf{p}_k \in \mathcal{P}_{\mathcal{S}}$, we can search for its nearby static points set $\mathcal{S}_{k}$ by radius search in Euclidean space. Let $|\mathcal{S}|$ be the cardinality of a set $\mathcal{S}$, the local smoothness is thus defined by: 
\begin{equation}
  \sigma_{k} = \frac{1}{|\mathcal{S}_{k}|}\cdot \sum_{\textbf{p}_{i} \in \mathcal{S}_{k}}(||\textbf{p}_{k}|| - ||\textbf{p}_{i}||).
\end{equation}
The edge features are defined by the points with large $\sigma_{k}$ and the planar features are defined by the points with small $\sigma_{k}$. 
\subsubsection{Data Association} The final robot pose is calculated by minimizing the point-to-edge and point-to-plane distance. For an edge feature point $\mathbf{p}_\mathcal{E} \in \mathbf{P}_\mathcal{E}$, it can be transformed into local map coordinate by $\hat{\mathbf{p}}_\mathcal{E} = \mathbf{T} \cdot \mathbf{p}_\mathcal{E}$, where $\mathbf{T} \in SE(3)$ is the current pose. We can search for 2 nearest edge features $\mathbf{p}_{\mathcal{E}}^1$ and $\mathbf{p}_{\mathcal{E}}^2$ from the local edge feature map and the point-to-edge residual is defined by \cite{wang2021light}: 
\begin{equation}
  f_{\mathcal{E}}(\hat{\mathbf{p}}_{\mathcal{E}}) = \frac{||(\hat{\mathbf{p}}_{\mathcal{E}} - \mathbf{p}_{\mathcal{E}}^1)\times (\hat{\textbf{p}}_{\mathcal{E}} - \mathbf{p}_{\mathcal{E}}^2)||}{||\mathbf{p}_{\mathcal{E}}^1-\mathbf{p}_{\mathcal{E}}^2||},
\end{equation}
where symbol $\times$ is the cross product. Similarly, given a planar feature point $\mathbf{p}_\mathcal{L} \in \mathcal{P}_\mathcal{L}$ and its transformed point $\hat{\mathbf{p}}_\mathcal{L} = \mathbf{T} \cdot \mathbf{p}_\mathcal{L}$, we can search for 3 nearest points $\textbf{p}_{\mathcal{L}}^1$, $\textbf{p}_{\mathcal{L}}^2$, and $\textbf{p}_{\mathcal{L}}^3$ from the local planar map. The point-to-plane residual is defined by: 
\begin{equation}
  f_{\mathcal{L}}(\hat{\textbf{p}}_\mathcal{L}) = (\hat{\textbf{p}}_{\mathcal{L}} - \textbf{p}_{\mathcal{L}}^1)^T \cdot \frac{(\textbf{p}_{\mathcal{L}}^1 - \textbf{p}_{\mathcal{L}}^2)\times (\textbf{p}_{\mathcal{L}}^1-\textbf{p}_{\mathcal{L}}^3)}{||(\textbf{p}_{\mathcal{L}}^1 - \textbf{p}_{\mathcal{L}}^2)\times (\textbf{p}_{\mathcal{L}}^1-\textbf{p}_{\mathcal{L}}^3)||}.
\end{equation}
\subsubsection{Pose Estimation} The final robot pose is calculated by minimizing the sum of point-to-plane and point-to-edge residuals:
\begin{equation}
   \textbf{T}^* = \text{arg}\min_{\textbf{T}}  \sum_{\mathbf{p}_\mathcal{E} \in \mathcal{P}_\mathcal{E}} f_{\mathcal{E}}(\hat{\textbf{p}}_\mathcal{E}) + \sum_{\mathbf{p}_\mathcal{L} \in \mathcal{P}_\mathcal{L}} f_{\mathcal{L}}(\hat{\textbf{p}}_\mathcal{L}).
\end{equation}
This non-linear optimization problem can be solved by the Gauss-Newton method and we can derive an optimal robot pose based on the static information.
\subsubsection{Feature Map Update \& Key Frame Selection}
Once the optimal pose is derived, the features are updated to the local edge map and local plane map respectively, which will be used for the data association on the next frame. Note that to build and update a global dense map is often very computational costly. Hence, the global static map is updated based on the keyframe. A key frame is selected when the translational change of the robot pose is greater than a predefined translation threshold, or the rotational change of the robot pose is greater than a predefined rotation threshold. 

\subsection{Global Map Building}
The semantic map is separated into a static map and a dynamic map. Note that the visual information given previously is also used to construct the colored dense static map. Specifically, the visual information can be achieved by re-projecting 3D points into the image plane. After each update, the map is down-sampled by using a 3D voxelized grid approach \cite{rusu20113d} in order to prevent memory overflow. The dynamic map is built by $\mathcal{P}_{\mathcal{D}}$ and it is used to reveal the dynamic objects. The dynamic information can be used for high-level tasks such as motion planning. 

\section{EXPERIMENT EVALUATION}\label{sec:experiment}

In this section, experimental results will be presented to demonstrate the effectiveness of our proposed method. First, our experimental setup will be discussed in detail. Second, we elaborate how we acquire the data of potential moving objects in a warehouse environment. Third, we evaluate the segmentation performance on our adopted instance segmentation model. Subsequently, we explain how we perform the dense mapping and dynamic tracking. Lastly, we evaluate the performance of our proposed method regarding the localization drifts under dynamic environments.
\subsection{Experimental Setup}
For our experimental setup, the Robot Operating System (ROS) is utilized as the interface for the integration of the semantic learning module and the SLAM algorithm, as shown in \fref{fig: title_graph}. Intel RealSense LiDAR camera L515 is used to capture RGB and point cloud at a fixed frame rate. All the experiments are performed on a computer with an Intel i7 CPU and an Nvidia GeForce RTX 2080 Ti GPU. 

\begin{figure}[t]
\begin{center}
\vspace{10pt}
\includegraphics[width=0.99\linewidth]{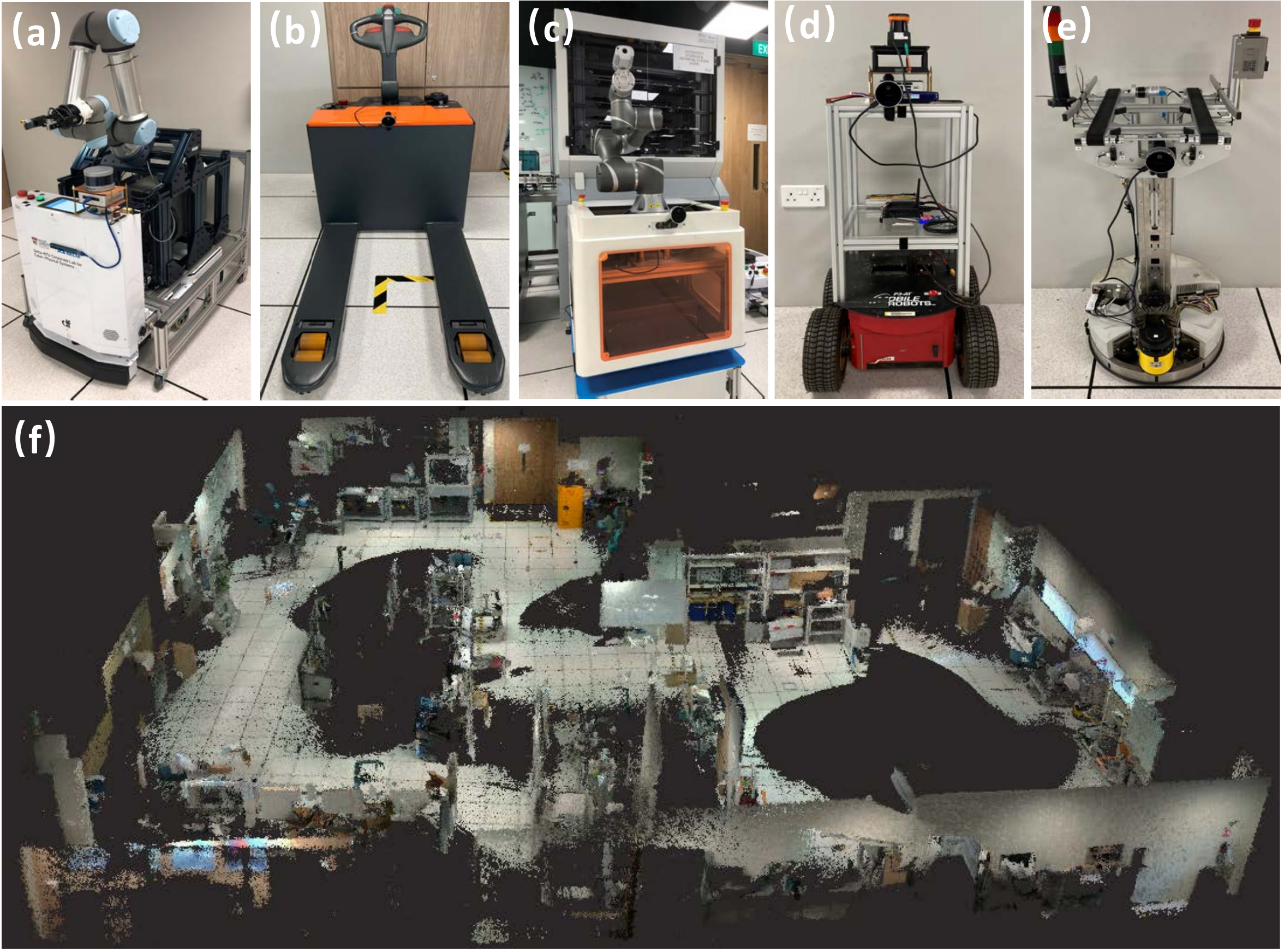}
\captionsetup{justification=justified}
\caption{Different types of AGVs used in our warehouse environment: (a) the grabbing AGV with a robot arm; (b) folklift AGV; (c) scanning AGV; (d) the Pioneer robot; (e) the transportation AGV with conveyor belt; (f) warehouse environment;}
\label{fig: agvs}
\end{center}
\end{figure}

\begin{figure*}[t] 
\begin{center}
\vspace{8pt}
\includegraphics[width=0.99\linewidth]{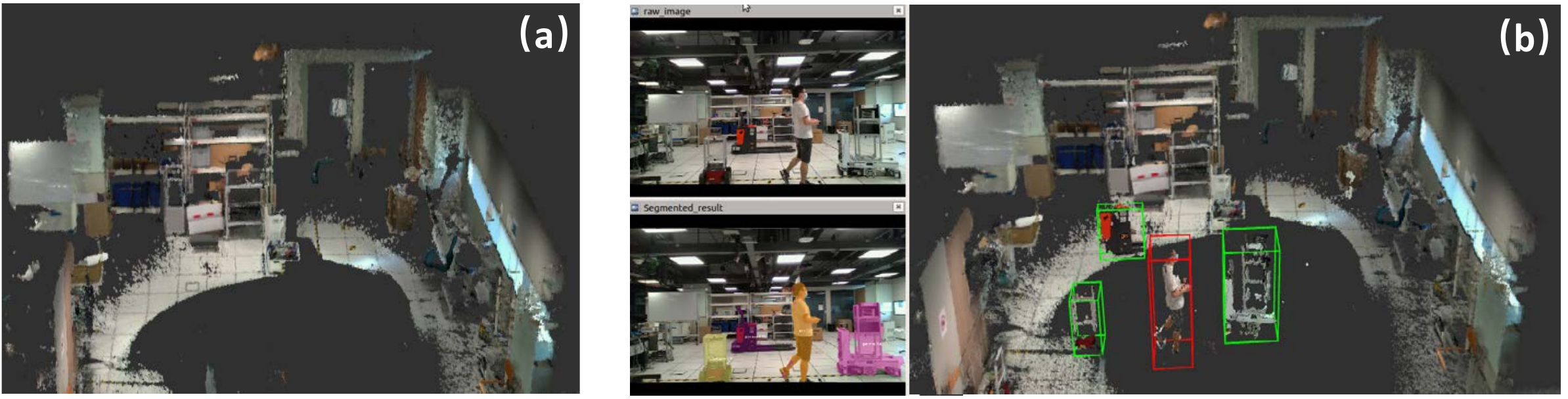}
\captionsetup{justification=justified}
\caption{Static map creation and final semantic mapping result: (a) static map built by the proposed SLAM framework; (b) final semantic mapping result. The instance segmentation is shown on the left. Human operators are labeled by red bounding boxes and AGVs are labeled by green bounding boxes.}
\label{fig: final_mapping}
\end{center}
\end{figure*}
 
\subsection{Data Acquisition}
Humans are often considered as dynamic objects in many scenarios such as autonomous driving and smart warehouse logistics. Therefore we choose 5,000 human images from the COCO dataset. In the experiment, the proposed method is evaluated in the warehouse environment as shown in \fref{fig: agvs}. Other than considering humans as dynamic objects, an advanced factory requires human-to-robot and robot-to-robot collaboration, so that the Automated Guided Vehicles (AGVs) are also potential dynamic objects. Hence a total of 3,000 AGV images are collected to train the instance segmentation network and some of the AGVs are shown in \fref{fig: agvs}. 

In order to solve the small dataset problem, we implement the copy-paste augmentation method proposed by \cite{ghiasi2020simple} to enhance the generalization ability of the network and directly improve the robustness of the network. To be specific, this method generates new images through applying random scale jittering on two random training datasets and randomly chooses a subset of object instances from one image to paste onto the other image.

\begin{table}[b]
    \begin{center}
    \setlength{\tabcolsep}{0.7\tabcolsep}
    \centering
    \begin{tabular}{cccc}
    \toprule
    \multirow{2}{*}{\textbf{Model}}   & \textbf{Segmentation} & \textbf{Mean} &\textbf{Inference}  \\
     & \textbf{Loss}& \textbf{AP (\%)}& \textbf{Time (ms)} \\
         \hline
    \midrule

    SOLOv2& 0.52& 38.8 &  \textbf{54.0} \\
    SOLOv2 + RFP& 0.36& 41.2 &64.0 \\
    SOLOv2 + SAC& 0.39& 39.8 &59.0 \\
    \textbf{SOLOv2+DetectoRS(Ours) }& \textbf{0.29} & \textbf{43.4} &71.0\\
    \bottomrule
    \end{tabular}
    \caption{Performance comparison of instance segmentation.}
    \label{table: performance comparison}
    \end{center}

\end{table}

\subsection{Evaluation on Instance Segmentation Performance}
In this part, we will evaluate the segmentation performance on the COCO dataset with regards to the segmentation loss and mean Average Precision (mAP). The purpose of this evaluation is to compare our adopted instance segmentation network, SOLOv2, with the proposed method. The results are illustrated in \tref{table: performance comparison}. Our adopted instance segmentation network, SOLOv2 is built based on the MMDetection 2.0 \cite{chen2019mmdetection}, an open-source object detection toolbox based on PyTorch. We trained SOLOv2 on the COCO dataset which consists of 81 classes. We choose ResNet-50 as our backbone architecture since this configuration satisfies our requirements for the real-world robotics applications. Instead of training the network from scratch, we make use of the parameters of ResNet-50 that are pre-trained on ImageNet. For fair comparison, all the models are trained under the same configurations, they are trained with the synchronized stochastic gradient descent with a total of 8 images per mini-batch for 36 epochs.

For SOLOv2 with Recursive Feature Pyramid (RFP), we modify our backbone architecture from Feature Pyramid Network (FPN) to RFP network. In this experiment, we only set the number of stages to be 2, allowing SOLOv2 to look at the image twice. As illustrated in \tref{table: performance comparison}, implementation of RFP network brings a significant improvement on the segmentation performance. On the other hand, we replace all 3x3 convolutional layers in the backbone network with Switchable Atrous Convolution (SAC), which increases the segmentation accuracy by 2.3\%. 
By implementing both SAC and RFP network to SOLOv2, the segmentation performance is further improved by 5.9\% with only 17\textit{ms} increase in inference time. Overall, SOLOv2 learns to look at the image twice with adaptive receptive fields, therefore it is able to highlight important semantic information for the instance segmentation network. The segmentation result is further visualized in \fref{fig: compare}.


\begin{table}[b]
    \begin{center}
    \begin{tabular}{ccc}
    \toprule
    \multirow{2}{*}{\textbf{Methods}}  & \textbf{ATDE} & \textbf{MTDE}  \\
    &\textbf{(cm)}& \textbf{(cm)}\\
    \hline
    \midrule
    W/O Semantic Recognition & 4.834    &   1.877 \\  
    Vision-based Semantic Recognition  & 1.273 & 0.667    \\  
    \textbf{Multi-Modal Recognition (Ours)}  &\textbf{ 0.875 }    &   \textbf{0.502} \\  
    \bottomrule
    \end{tabular}
    \caption{Ablation study of localization drifts under dynamic environments.}
    \label{table:ablation study}
    \end{center}
\end{table}




\subsection{Dense Mapping and Dynamic Tracking}
\begin{figure}[t]
\begin{center}
\vspace{0pt}
\includegraphics[width=0.89\linewidth]{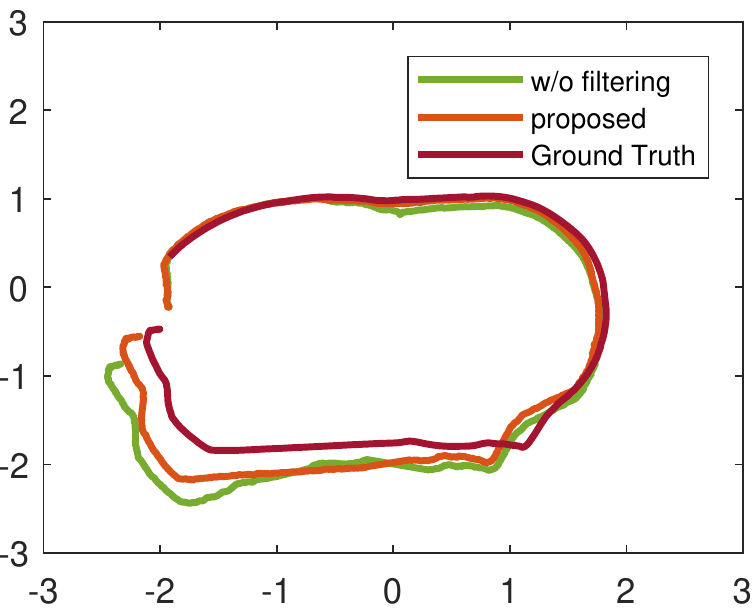}
\captionsetup{justification=justified}
\caption{Localization comparison in a dynamic environment. The ground truth, the original localization result without filtering and the localization result with our proposed multi-modal semantic filtering are plotted in red, green and orange respectively. }
\label{fig: vicon}
\end{center}
\end{figure}

To evaluate the performance of our multi-modal semantic SLAM in dynamic environments, the proposed method is implemented on warehouse AGVs which are shown in \fref{fig: agvs}. In a smart manufacturing factory, both human operators and different types of AGVs (e.g., folklift AGVs, transportation AGVs and robot-arm equipped AGVs) are supposed to work in a collaborative manner. Therefore, the capability of each AGV to localize itself under moving human operators and other AGVs is the essential technology towards industry 4.0. 
In many warehouse environments, the rest of objects such as operating machines or tables can be taken as a static environment. Hence we only consider humans and AGVs as dynamic objects in order to reduce the computational cost. In the experiment, an AGV is manually controlled to move around and build the warehouse environment map simultaneously, while the human operators are walking frequently in the warehouse. The localization result is shown in \fref{fig: vicon}, where we compare the results of ground truth, the proposed SLAM method and original SLAM without our filtering approach. It can be seen that when the dynamic object appears (in blue), the proposed multi-modal semantic SLAM is more robust and stable than traditional SLAM. The mapping results are shown in \fref{fig: final_mapping}. The proposed method is able to efficiently identify the potential dynamic objects and separate them from the static map. Although the human operators are walking frequently in front of the robot, they are totally removed from the static map. All potential dynamic objects are enclosed by bounding boxes and are added into a final semantic map to visualize the status of each object in real time, where the moving human is colored in red and the AGVs are colored in green. Our method is able to identify and locate multiple targets in the complex dynamic environment.

\begin{figure}[t]
\begin{center}
\vspace{0pt}
\includegraphics[width=0.99\linewidth]{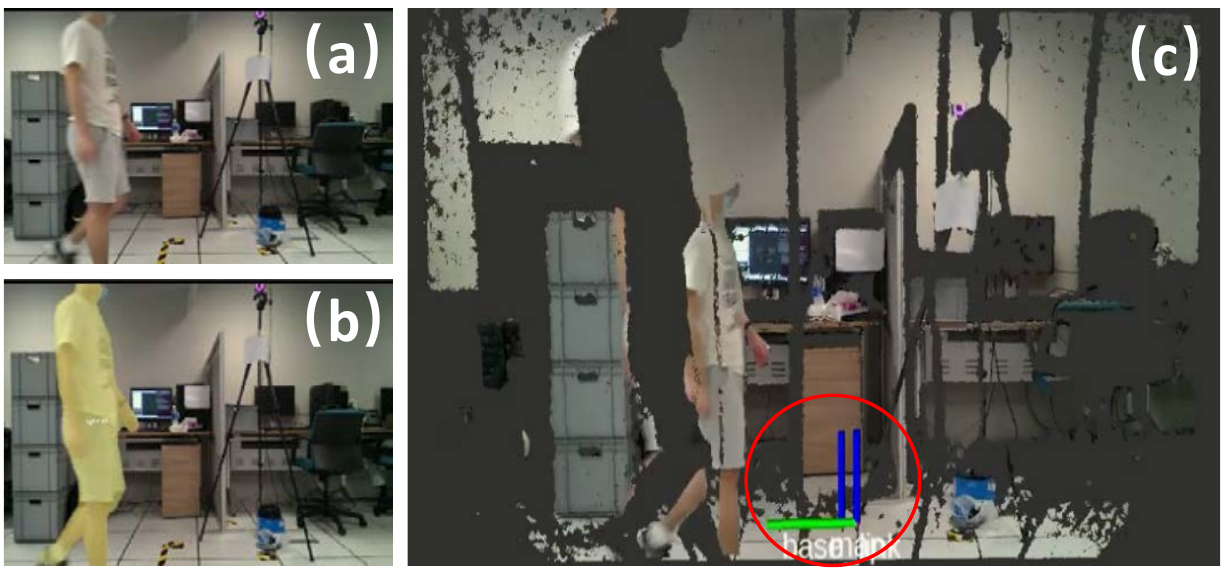}
\captionsetup{justification=justified}
\caption{Ablation study of localization drifts. (a) original image view; (b) the visual semantic recognition result based on the proposed method; (c) Localization drifts observed due to the moving objects. The localization drifts are highlighted in red circle. }
\label{fig: localization_drift}
\end{center}
\end{figure}

\subsection{Ablation Study of Localization Drifts}
To further evaluate the performance of localization under dynamic profiles, we compare the localization drifts of different dynamic filtering approaches. Firstly, we keep the robot still and let a human operator walk frequently in front of the robot. The localization drifts are recorded in order to evaluate the performance under dynamic objects. Specifically, we calculate the Average Translational Drifts Error (ATDE) and Maximum Translational Drifts Error (MTDE) to verify the localization, where the ATDE is the average translational error of each frame and MTDE is the maximum translational drift caused by the walking human. The results are shown in \tref{table:ablation study}. We firstly remove the semantic recognition module from SLAM and evaluate the performance. Then we use the visual semantic recognition (SOLOv2) to remove the dynamic information. The results are compared with the proposed semantic multi-modal SLAM. It can be seen that, compared to the original SLAM, the proposed method significantly reduces the localization drift. Compared to vision-only-based filtering methods, the proposed multi-modal semantic SLAM is more stable and accurate under the presence of dynamic objects.

\section{CONCLUSION}\label{sec:conclusion}
In this paper, we have presented a semantic multi-modal framework to tackle the SLAM problem in dynamic environments, which is able to effectively reduce the impact of dynamic objects in complex dynamic environments. Our approach aims to provide a modular pipeline to allow real-world applications in dynamic environments. Meanwhile, a 3D dense stationary map is constructed with the removal of dynamic information. To verify the effectiveness of the proposed method in a dynamic complex environment, our method is evaluated on warehouse AGVs used for smart manufacturing. The results show that our proposed method can significantly improve the existing semantic SLAM algorithm in terms of robustness and accuracy. 


\balance
\bibliographystyle{IEEEtran}
\bibliography{IEEEabrv,references}

\end{document}